\def\CameraReadyClean{1}
\def\CostgradPreprint{1}
\PassOptionsToPackage{final}{neurips_2026}
\documentclass{article}

\usepackage[main,nonanonymous]{neurips_2026}

\usepackage[utf8]{inputenc}
\usepackage[T1]{fontenc}
\usepackage{hyperref}
\usepackage{url}
\usepackage{booktabs}
\usepackage{amsfonts}
\usepackage{amsmath}
\usepackage{amssymb}
\usepackage{bm}
\usepackage{graphicx}
\usepackage{placeins}
\usepackage{multirow}
\usepackage{nicefrac}
\usepackage{microtype}
\usepackage{xcolor}

\ifdefined\CostgradPreprint
\definecolor{costgradlink}{RGB}{25,70,130}
\hypersetup{colorlinks=true,allcolors=costgradlink,pdfborder={0 0 0}}
\fi

\newif\ifrevisionmarks
\ifdefined\CameraReadyClean\revisionmarksfalse\else\revisionmarkstrue\fi
\newcommand{\rev}[1]{{\ifrevisionmarks\color{red}\fi #1}}
\newenvironment{revision}{\begingroup\ifrevisionmarks\color{red}\fi}{\endgroup}

\ifdefined\CostgradPreprint\else
\makeatletter
\renewcommand{\@noticestring}{\rev{\@trackname}}
\makeatother
\fi

\usepackage{subcaption}
\usepackage{wrapfig}
\usepackage{natbib}

\usepackage{float}
\floatstyle{ruled}
\newfloat{algorithm}{tbp}{loa}
\floatname{algorithm}{Algorithm}
\newcounter{algoline}
\renewcommand{\thealgoline}{\arabic{algoline}}
\newenvironment{algorithmic}[1][]{%
  \setcounter{algoline}{0}%
  \begin{list}{\refstepcounter{algoline}\thealgoline:}{%
    \setlength{\labelwidth}{2em}%
    \setlength{\leftmargin}{2em}%
    \setlength{\itemsep}{1pt}%
    \setlength{\parskip}{0pt}%
    \small%
  }%
}{\end{list}}
\newcommand{\State}{\item}
\newcommand{\Comment}[1]{\hfill\textit{// #1}}
\newcommand{\Return}{\textbf{return}}

\newcommand{\costgrad}{\textsc{CostGrad}}
\newcommand{\predattn}{\textsc{PredAttn}}
\newcommand{\predgrad}{\textsc{PredGrad}}
\newcommand{\pp}{\,\text{pp}}
\newcommand{\reals}{\mathbb{R}}

\title{Sparse Planning in Visual World Models \\ via Cost Gradients}

\author{%
  \rev{Yingchen Xu} \\
  \rev{University College London} \\
  \rev{\texttt{yingchen.xu.21@ucl.ac.uk}}
  \And
  \rev{Edward Grefenstette} \\
  \rev{University College London} \\
  \rev{\texttt{e.grefenstette@ucl.ac.uk}}
}

\graphicspath{{./}}

\begin{document}
\maketitle
\raggedbottom

\begin{abstract}
\rev{Token-based world models enable fine-grained latent planning, but repeatedly processing large spatial token grids makes action search expensive.}
We introduce \costgrad{}, a training-free, goal-conditioned selector that ranks spatial tokens by the gradient norm of the planning cost with respect to each input token.
By deriving importance from the downstream control objective, \costgrad{} targets tokens that matter for planning rather than merely for prediction.
On AdaLN-conditioned predictors at $50\%$ sparsity, \costgrad{} matches or exceeds full-token planning on three of four continuous-control benchmarks, while giving a measured $2.6\times$ wall-clock speedup per environment planning step.
\rev{Combining token sparsity with reduced CEM search increases this to a $\sim\!5\times$ total speedup while still exceeding the full-token baseline.}
We also identify an architecture-dependent failure mode: in a matched AdaLN-vs-concat comparison, concat maintains comparable full-token performance but pure \costgrad{} loses its advantage over random selection.
\rev{This difference tracks action-pathway drift: gradient-selected removal produces less drift than random removal on AdaLN, but more on concat.}
\rev{These results highlight selector--architecture compatibility as a design axis for sparse world-model planning.}
\rev{Project page and demos: \href{https://ycxuyingchen.github.io/costgrad/}{\nolinkurl{ycxuyingchen.github.io/costgrad/}}.}
\end{abstract}

\section{Introduction}\label{sec:intro}

Token-based world models~\citep{zhou2025dinowm,terver2025jepawms} \rev{predict future spatial feature grids produced by a frozen vision encoder}. At inference time, planning runs the cross-entropy method (CEM)~\citep{rubinstein1999cem} inside the learned model, rolling out many candidate action sequences against an encoded goal. This is expensive: at $16 \times 16$ tokens, horizon $6$, and standard CEM widths, one planning step performs $\sim\!10^7$ token-level forward computations. Much of this compute is often spent on weakly task-relevant tokens: in control tasks, the agent, object, obstacle, and goal occupy only a small part of the frame, while most patches describe background. \rev{Sparse Imagination~\citep{chun2025sparseimagination} demonstrates efficient world-model planning through sparse training and random token selection. We ask how a fixed pretrained predictor can select tokens relevant to the current goal without sacrificing planning success.}

A natural approach is to reuse the predictor's own signals, such as attention weights or prediction error. \rev{But prediction is not planning: improving dynamics prediction does not necessarily improve downstream control, an objective mismatch documented in model-based reinforcement learning~\citep{lambert2020objective}.} Attention can be underdetermined by prediction loss, and prediction error highlights visual unpredictability rather than control relevance: the object most important for control may be easy to predict because its motion follows directly from the action. This motivates using the planning objective itself as the source of token importance.

We instantiate this idea with \emph{\costgrad{}}: given an encoded observation and goal, we compute the gradient of the planning cost with respect to each input token, and rank tokens by gradient norm.
\rev{Subsequent action search uses only the selected tokens.}
\rev{\costgrad{} requires one extra forward and backward pass per planning step, operates on a fixed pretrained predictor without retraining, and conditions selection on the current goal.}
On AdaLN-conditioned predictors at $50\%$ sparsity, \costgrad{} matches or exceeds full-token planning success on three of four environments \rev{and gives} a measured $2.6\times$ wall-clock speedup per environment planning step.

The more surprising finding is that \costgrad{} depends on how actions enter the predictor. We compare two otherwise matched predictors: AdaLN~\citep{peebles2023dit} injects actions by modulating each visual token with action-dependent parameters, while concat appends the action to each token before self-attention. Pure \costgrad{} succeeds on AdaLN but loses its advantage on matched-concat, despite comparable full-token planning. \rev{This difference tracks action-pathway drift: gradient-selected token removal produces less drift than random removal on AdaLN, but more on concat.} Sparse planning therefore depends on both choosing relevant tokens and on a predictor architecture that remains stable when restricted to those tokens. More generally, the planner is part of the model's operating distribution: token selection changes not just what the model sees, but how the planner uses the model.

\textbf{Contributions.}
(i) We introduce \costgrad{}, a training-free, goal-conditioned token selector for sparse visual world-model planning. At $50\%$ sparsity, \costgrad{} matches or exceeds full-token planning on three of four benchmarks, gives a measured $2.6\times$ wall-clock speedup per environment planning step, and can combine token sparsity with reduced CEM iterations for a $\sim\!5\times$ total speedup while still exceeding the full-token baseline (\S\ref{sec:results_main}).
(ii) We identify a selector--architecture compatibility failure mode. \rev{In a matched AdaLN-vs-concat comparison, \costgrad{} beats random selection on AdaLN but loses its advantage on concat. Gradient-selected removal reduces action-pathway drift relative to random removal on AdaLN but increases it on concat.} Anchor-mixing experiments further show that concat can require more coverage-preserving sparse supports, suggesting that sparse planning depends on matching the selector to the predictor architecture (\S\ref{sec:swap}).

\section{Setup and Method}
\label{sec:method}

\subsection{Token-based world model planning}
\label{sec:setup}

We build on the DINO-WM framework~\citep{zhou2025dinowm} using the open-source JEPA-WMs codebase~\citep{terver2025jepawms}.
A frozen DINOv2 ViT-S/14 encoder~\citep{oquab2024dinov2} maps each $224 \times 224$ image to a $16 \times 16$ grid of $384$-dimensional tokens, $\bm{z} \in \reals^{P \times D}$ with $P = 256$.
\rev{Our main results use an AdaLN-Zero-conditioned predictor~\citep{peebles2023dit}: a $6$-block transformer with $16$ attention heads and GELU activation that maps the encoded context and an action to the next-step token grid.
It is trained on trajectory data from \citet{terver2025jepawms} and kept fixed during planning; the block equations are given in Appendix~\ref{app:invariance}.}

\begin{revision}
Planning uses CEM~\citep{rubinstein1999cem} in latent space: at each environment step, CEM samples $300$ action sequences over horizon $H{=}6$, ranks them by cost to the encoded goal, and refits the sampling distribution to the top elites for $30$ iterations (JEPA-WMs defaults).
Each candidate is rolled out autoregressively, calling the predictor $H{=}6$ times on the full $P{=}256$-token grid, so one planning step performs $300 \times 30 \times 6 \times 256 \approx 1.4 \times 10^7$ token-level forward computations.
\end{revision}

\subsection{\rev{\costgrad{}: select tokens, then plan}}
\label{sec:costgrad}

\begin{figure}[!t]
\ifrevisionmarks\color{red}\captionsetup{labelfont={color=red}}\fi
\centering
\includegraphics[width=\linewidth]{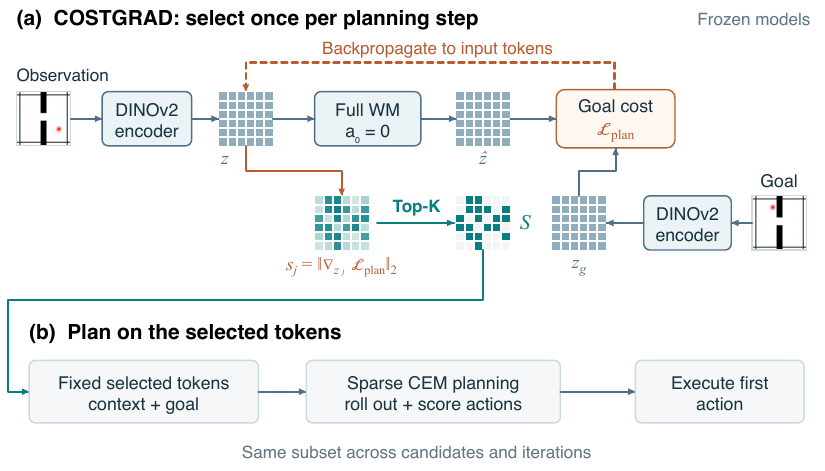}
\caption{\rev{\textbf{\costgrad{}: select once, then plan on the selected tokens.}
\textbf{(a)} A full-token, zero-action probe supplies planning-cost gradients to select top-$K$ positions $\mathcal S$ for context and goal.
\textbf{(b)} CEM searches for actions using only $\mathcal S$, then executes the first action.
The subset stays fixed throughout action search and is reselected at the next observation; model weights remain frozen.
Token grids are schematic; proprioception is omitted.}}
\label{fig:method_overview}
\end{figure}

\costgrad{} scores tokens using the \emph{planning} objective rather than the predictor's training objective.
\rev{Figure~\ref{fig:method_overview} illustrates how a single selection probe defines the token subset reused throughout action search.}

\paragraph{\rev{Select tokens.}}
\rev{Given a context encoding $\bm{z}$ and goal encoding $\bm{z}_g$, we run a one-step prediction on all $P$ tokens under a probe action $\bm{a}_0$ (zero by default) and compute}
\begin{equation}
  \mathcal{L}_{\text{plan}}(\bm{a}_0; \bm{z}, \bm{z}_g) = \big\| \mathrm{predictor}(\bm{z}, \bm{a}_0) - \bm{z}_g \big\|_2^2\rev{.}
\end{equation}
\rev{A single backward pass gives the per-token score}
\begin{equation}
  s_j = \big\| \nabla_{\bm{z}_j} \mathcal{L}_{\text{plan}} \big\|_2 ,\qquad j = 1, \dots, P.
\end{equation}
\rev{We retain the $K$ tokens with the largest scores, defining the selected set $\mathcal{S}$.}

\begin{revision}
The gradient norm measures local planning-cost sensitivity to each input token, including its influence on predictions elsewhere.
Thus, a token can receive a high score by providing context for predicting goal-relevant regions, even when its own predicted feature is close to the goal.
\end{revision}

\begin{revision}
\paragraph{Plan on the selected tokens.}
CEM rolls out each candidate action sequence autoregressively for the full horizon $H$ using only $\mathcal{S}$ and evaluates the final goal-distance cost at the same kept positions.
The subset remains fixed across all candidates and CEM iterations; we reselect tokens at the next observation.
The one-step probe is used only for selection, while CEM samples and optimizes its own actions (Algorithm~\ref{alg:costgrad}).
We retain CEM to match the JEPA-WMs evaluation protocol and isolate the effect of token selection.
\end{revision}

\begin{revision}
\paragraph{Implementation.}
Retained tokens keep their original positional indices, shared across context frames and the goal.
Gradients score the latest observed frame; rollout uses up to two context frames.
The probe uses only visual squared error and zero normalized action/proprioceptive inputs.
CEM combines final-step mean squared visual and proprioceptive-feature errors with weights $1$ and $0.1$, respectively (objective and normalization in Appendix~\ref{app:hyperparams}).
\end{revision}

\begin{revision}
\paragraph{Why the planning objective?}
Prediction error measures what the model gets wrong, whereas planning-cost gradients measure how its inputs affect the goal-reaching objective.
Even a perfectly predicted next state can remain far from the goal, so prediction accuracy alone does not identify which tokens matter for planning.
Appendix~\ref{app:planning_preservation} develops this objective distinction and the local-sensitivity interpretation.
\end{revision}

\begin{algorithm}[t]
\caption{\costgrad{} sparse planning step}
\label{alg:costgrad}
\begin{algorithmic}[1]
\State \textbf{Input:} encoded context $\bm{z}$, goal $\bm{z}_g$, keep ratio $K/P$
\State \textit{// Step 1: select tokens once using a probe action} $\bm{a}_0$
\State Set probe action $\bm{a}_0 \gets \bm{0}$ \Comment{\textit{zero-action by default; Appendix~\ref{app:probe} for alternatives}}
\State $\hat{\bm{z}} \gets \mathrm{predictor}(\bm{z}, \bm{a}_0)$ \Comment{\textit{full-token forward, retain graph}}
\State $\mathcal{L} \gets \| \hat{\bm{z}} - \bm{z}_g \|_2^2$
\State $s_j \gets \| \nabla_{\bm{z}_j} \mathcal{L} \|_2$ for all tokens $j$ \Comment{\textit{single backward pass}}
\State $\mathcal{S} \gets \mathrm{top}\text{-}K(s_j)$ \Comment{\textit{sparse token set, fixed for the rest of this planning step}}
\State \textit{// Step 2: run CEM as usual on the sparse token set; CEM samples its own actions}
\State \rev{\Return\ $\mathrm{CEM}(\bm{z}_{\mathcal{S}}, \bm{z}_{g, \mathcal{S}};\,\text{horizon } H, 30 \text{ iters}, 300 \text{ candidates})$}
\end{algorithmic}
\end{algorithm}

\begin{revision}
\paragraph{Computational cost.}
\costgrad{} is training-free and goal-conditioned: selection depends on the current goal while predictor weights remain fixed.
Its overhead is one full-token forward and backward pass per planning step, amortized over all CEM rollouts.
Attention compute scales quadratically in token count while MLP scales linearly, so halving tokens reduces attention flops by $4\times$ and MLP flops by $2\times$.
\end{revision}

\newpage
\begin{samepage}
\subsection{Baselines}
\label{sec:baselines}

\rev{We compare \costgrad{} with \textbf{Full} and four token-reduction baselines, sharing the trained predictor and CEM settings within each environment.
Full retains all $P=256$ tokens; all token-reduction methods use $K=P/2=128$.}
\par\end{samepage}

\begin{revision}
\textbf{\predattn{}} ranks spatial keys by received attention in the predictor's last block under a full-grid, zero-action probe: sum over all queries and average uniformly over heads and context frames, with causal masking and conditioning-token keys excluded.

\textbf{\predgrad{}} ranks tokens by prediction-loss gradient norms, using earlier observed frames as input and the latest observed encoding as $\bm z_{\mathrm{target}}$, without privileged future information.
It uses a zero-action probe; the observed target need not match that action.
Appendix~\ref{app:hyperparams} specifies the loss and temporal alignment.

\textbf{Random} samples $K$ positions uniformly without replacement, once per planning step, and fixes them throughout CEM.

\textbf{ToMe}~\citep{bolya2023tome} merges observation tokens by cosine-based bipartite matching, averaging each group at its surviving position.
The mapping is shared across context frames; rollout predicts at the surviving positions and compares against unmerged goal features there (Appendix~\ref{app:hyperparams}).

\end{revision}

\suppressfloats[t]
\newlength{\resultsFloatSep}
\setlength{\resultsFloatSep}{\textfloatsep}
\setlength{\textfloatsep}{12pt plus 2pt minus 2pt}
\section{Results}
\label{sec:results}

\rev{We evaluate planning performance and speed (\S\ref{sec:results_main}), the task relevance and stability of selected tokens (\S\ref{sec:why_planning}), and the dependence of sparse planning on predictor architecture (\S\ref{sec:swap}).}

\begin{revision}
We evaluate on four continuous-control benchmarks from the DINO-WM and JEPA-WMs suites --- \textbf{PointMaze (MZ)}, \textbf{Wall}, \textbf{PushT (PT)}, and \textbf{MetaWorld (MW)} --- detailed in Appendix~\ref{app:envs}.
\end{revision}

\pagebreak[0]
\begin{samepage}
\subsection{Sparse planning performance and speed on AdaLN-Zero}
\label{sec:results_main}

\begin{table}[t]
\centering
\small
\caption{\textbf{Main results at $50\%$ tokens.}
Planning success rate ($\%$) across four continuous-control environments.
Mean $\pm$ std over 3 evaluation seeds (96 episodes each).
Full uses all $256$ tokens; all other methods keep $128$ tokens. \textbf{Bold}: best non-Full method per environment.
\costgrad{} preserves or exceeds Full on three of four environments while giving a measured $2.6\times$ wall-clock speedup per environment planning step.}
\label{tab:main}
\begin{tabular}{lccccc}
\toprule
Method ($50\%$) & PointMaze (MZ) & Wall & PushT (PT) & MetaWorld (MW) & Avg \\
\midrule
Full ($100\%$)   & $88.2 \pm 8.7$ & $84.7 \pm 2.6$ & $62.4 \pm 4.4$ & $58.7 \pm 1.8$ & $73.5$ \\
\midrule
\textbf{\costgrad{}} & $86.2 \pm 6.1$ & $\mathbf{92.0 \pm 1.8}$ & $\mathbf{65.2 \pm 6.5}$ & $\mathbf{65.6 \pm 4.7}$ & $\mathbf{77.3}$ \\
\predattn{}      & $\mathbf{86.7 \pm 9.7}$ & $67.3 \pm 1.7$ & $62.0 \pm 8.0$ & $44.8 \pm 6.1$ & $65.2$ \\
\predgrad{}       & $76.8 \pm 9.2$ & $67.2 \pm 3.3$ & $41.3 \pm 5.9$ & $51.4 \pm 3.5$ & $59.2$ \\
ToMe              & $74.4 \pm 9.1$ & $84.4 \pm 2.6$ & $51.4 \pm 5.9$ & $39.1 \pm 4.1$ & $62.3$ \\
Random            & $77.0 \pm 9.4$ & $67.0 \pm 3.6$ & $41.3 \pm 5.9$ & $46.9 \pm 3.1$ & $58.1$ \\
\bottomrule
\end{tabular}
\end{table}

Table~\ref{tab:main} reports planning success at $K{=}128$ tokens.
\costgrad{} preserves full-token performance on PointMaze ($-2.0\pp$, within noise) and exceeds Full on Wall ($+7.3\pp$), PushT ($+2.8\pp$), and MetaWorld ($+6.9\pp$), giving an average gain of $+3.8\pp$ while using half the tokens.
Prediction-derived selectors are substantially weaker: the closest baseline, \predattn{}, trails \costgrad{} by $12.1\pp$ on average and by more than $20\pp$ on Wall and MetaWorld.
Random selection is far worse ($-19.2\pp$ on average), showing that the gain on AdaLN-Zero is not a generic sparsification effect; \S\ref{sec:swap} examines what changes when the predictor's action-conditioning pathway is replaced.
\par\end{samepage}

\rev{On released JEPA-WMs checkpoints, \costgrad{} also preserves full-token performance across all four environments without retraining, using half the tokens and outperforming Random by $11.3$--$46.3\pp$ (Appendix~\ref{app:dropin}, Table~\ref{tab:dropin}).}

\begin{revision}
\begin{figure}[!b]
\ifrevisionmarks\color{red}\captionsetup{labelfont={color=red}}\fi
\centering
\includegraphics[width=\linewidth]{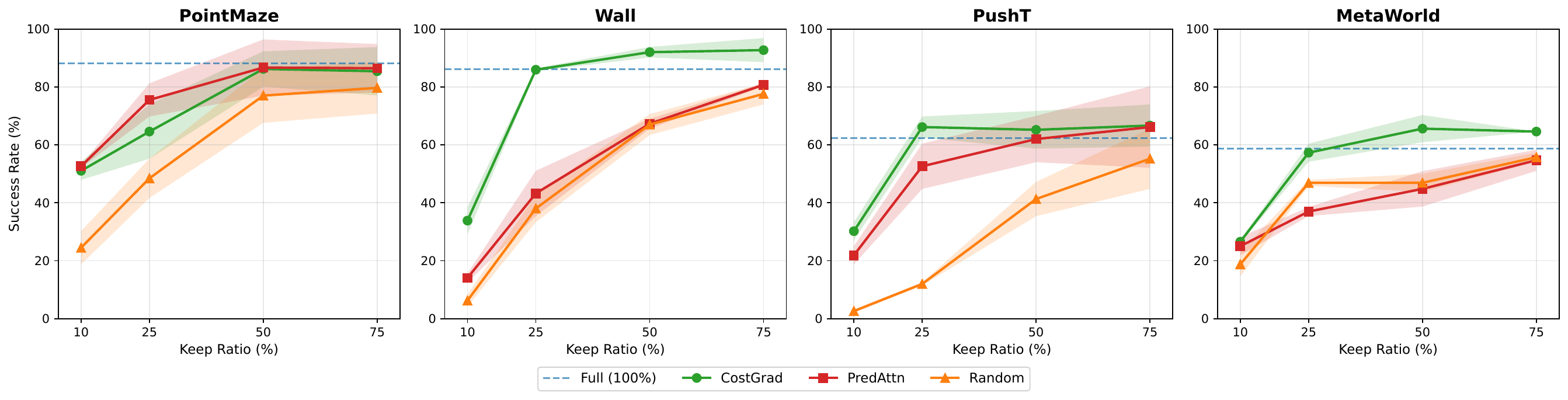}
\caption{\rev{\textbf{Sparsity sweeps on four environments.} Planning success ($\%$) vs.\ keep ratio for \costgrad{}, \predattn{}, and Random; horizontal dashed line is Full ($100\%$). Shaded bands: $\pm 1$ std across $3$ evaluation seeds at keep ratios $10$/$25$/$75$ ($96$ episodes per cell per seed); the $50\%$ point uses Table~\ref{tab:main}'s 3-eval-seed mean. \costgrad{} approaches or exceeds Full at $25\%$ tokens on three of four environments (Wall, PushT, MetaWorld); the gap to prediction-derived baselines (\predattn{}, Random) widens as sparsity increases. All sparse methods degrade sharply below $\sim\!25\%$.}}
\label{fig:sparsity}
\end{figure}

\pagebreak[0]
\begin{samepage}
\paragraph{Sparsity sweep.}
Figure~\ref{fig:sparsity} varies the keep ratio from $10\%$ to $100\%$.
\costgrad{} remains near Full at $25\%$ tokens on Wall, PushT, and MetaWorld, and approaches Full at $50\%$ on PointMaze.
Below roughly $25\%$, all sparse methods degrade sharply, indicating that the selected subset must still contain enough state information for rollout.
The gap to \predattn{} and Random widens as the token budget shrinks: planning-gradient selection becomes more important in the high-sparsity regime.
\par\end{samepage}
\end{revision}
\setlength{\textfloatsep}{\resultsFloatSep}

\begin{revision}
\pagebreak[0]
\begin{samepage}
\paragraph{\costgrad{} dominates the measured speed--success frontier.}
Retaining $50\%$ of the tokens with the standard CEM budget gives the $2.6\times$ speedup reported in Table~\ref{tab:main}.
The extra forward and backward pass adds $27.5$~ms on a single H100, amortized over the subsequent CEM iterations.

Token selection and CEM budget are complementary compute axes: predictor cost per CEM iteration scales with the number of kept tokens $K$, and total planning-step time scales linearly with the number of CEM iterations.
Combining $50\%$ tokens with $15$ CEM iterations reaches $92.0\%$ success at $\sim\!17$~s per planning step --- a $5.2\times$ total speedup over Full ($84.7\%$ at $88.6$~s) at higher mean success (Appendix~\ref{app:matched_compute}).
At the same $15$-iteration budget, Full achieves $84.4\pm1.0\%$ success, whereas \costgrad{} achieves $92.0\pm5.1\%$ while using half the tokens (Appendix~\ref{app:matched_compute}).
\par\end{samepage}

Across the measured token budgets and CEM settings (Figure~\ref{fig:pareto}), \costgrad{} configurations trace the entire non-dominated frontier; Full, \predattn{}, and Random are dominated.
For example, \costgrad{} at $K{=}192$ reaches $92.7\%$ success at $60.4$~s per step ($+8.0\pp$ and $1.47\times$ faster than Full), while the \predattn{} and Random sweep curves trail \costgrad{} by large margins.
Success enters a high-performance plateau around $K \ge 64$ across the measured \costgrad{} configurations, with the best points near $\sim\!92\%$.

\begin{figure}[!tb]
\ifrevisionmarks\color{red}\captionsetup{labelfont={color=red}}\fi
\centering
\includegraphics[width=\linewidth]{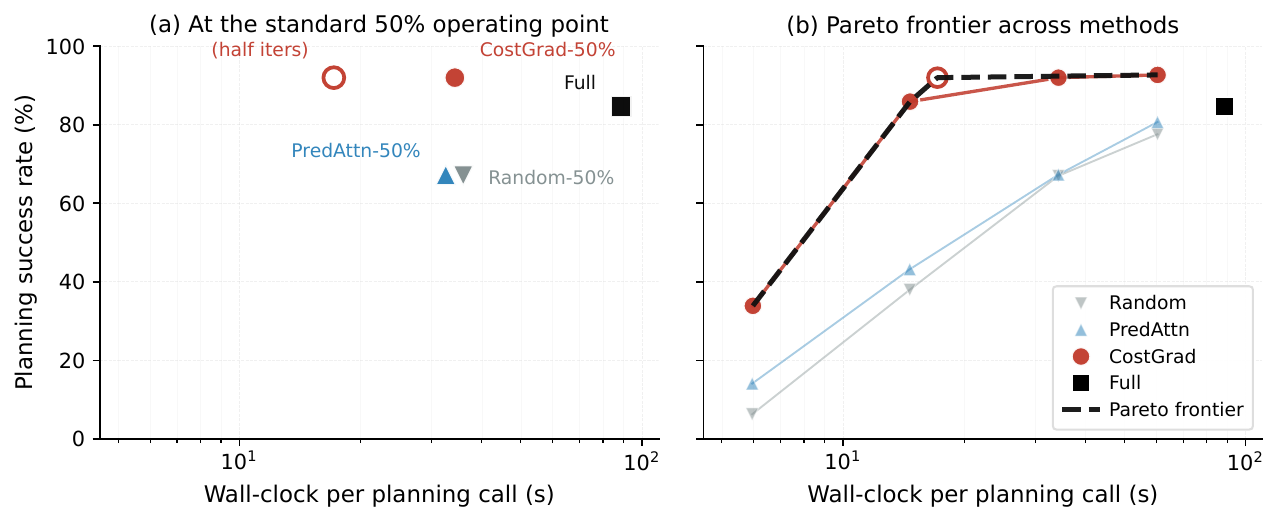}
\caption{\rev{\textbf{Wall speed--success trade-off on AdaLN-Zero.}
\textbf{(a)} Five canonical operating points at the standard $K{=}128$ sparse setting. \costgrad{}-50\% with $15$ CEM iterations achieves the same mean success as standard \costgrad{}-50\% while reducing planning time to $\sim\!17$~s, a $5.2\times$ speedup over Full.
\textbf{(b)} Pareto frontier across methods at $30$ CEM iterations, plus the \costgrad{} $K{=}128$, $15$-iteration point. Among measured points, the non-dominated frontier is traced by \costgrad{} configurations; Full, \predattn{}, and Random are dominated.
Timing details and the full $(K, \text{iters})$ marker convention are in Appendix~\ref{app:timing} and Appendix~\ref{app:cg_curve}.}}
\label{fig:pareto}
\end{figure}
\end{revision}

\begin{revision}
\pagebreak[0]
\begin{samepage}
\paragraph{Why can sparsity exceed Full?}
Sparse planning also changes the effective planning cost: Full includes background and weakly task-relevant tokens that may add noise to the objective.
Restricting rollout and goal comparison to high-sensitivity tokens can sharpen the cost landscape around controllable objects and goal-relevant regions, provided the subset retains enough state for prediction.
This interpretation is consistent with the cost-function ablation on Wall: \costgrad{}'s gain over Full \emph{triples} when the cost is more degenerate --- $+22.9\pp$ under L$_1$ versus $+7.3\pp$ under L$_2$ (Appendix~\ref{app:cost_function}).
\par\end{samepage}
\end{revision}

\pagebreak[0]
\begin{samepage}
\subsection{\rev{Task relevance and stability of planning-gradient selection}}
\label{sec:why_planning}

\S\ref{sec:results_main} showed that planning-gradient selection outperforms attention- and prediction-error-based selection by large margins on AdaLN-Zero (Table~\ref{tab:main}).
\rev{We first examine what \costgrad{} selects and how those selections align with task relevance, then test their stability across trained predictors and probe actions.}

\begin{revision}
\paragraph{Goal-conditioned selection.}
Figure~\ref{fig:cg_envs} visualizes \costgrad{} selections across environments.
The kept tokens concentrate on task-relevant regions --- maze corridors, the agent and wall, the pusher and block, the robot arm and object --- while dropping much of the background.
Decoded one-step predictions from the same sparse token sets remain qualitatively faithful (Appendix~\ref{app:decoded_pred}, Figure~\ref{fig:cg_envs_decoded}).
\end{revision}
\par\end{samepage}

\begin{revision}
\begin{figure}[!tb]
\ifrevisionmarks\color{red}\captionsetup{labelfont={color=red}}\fi
\centering
\includegraphics[width=\linewidth]{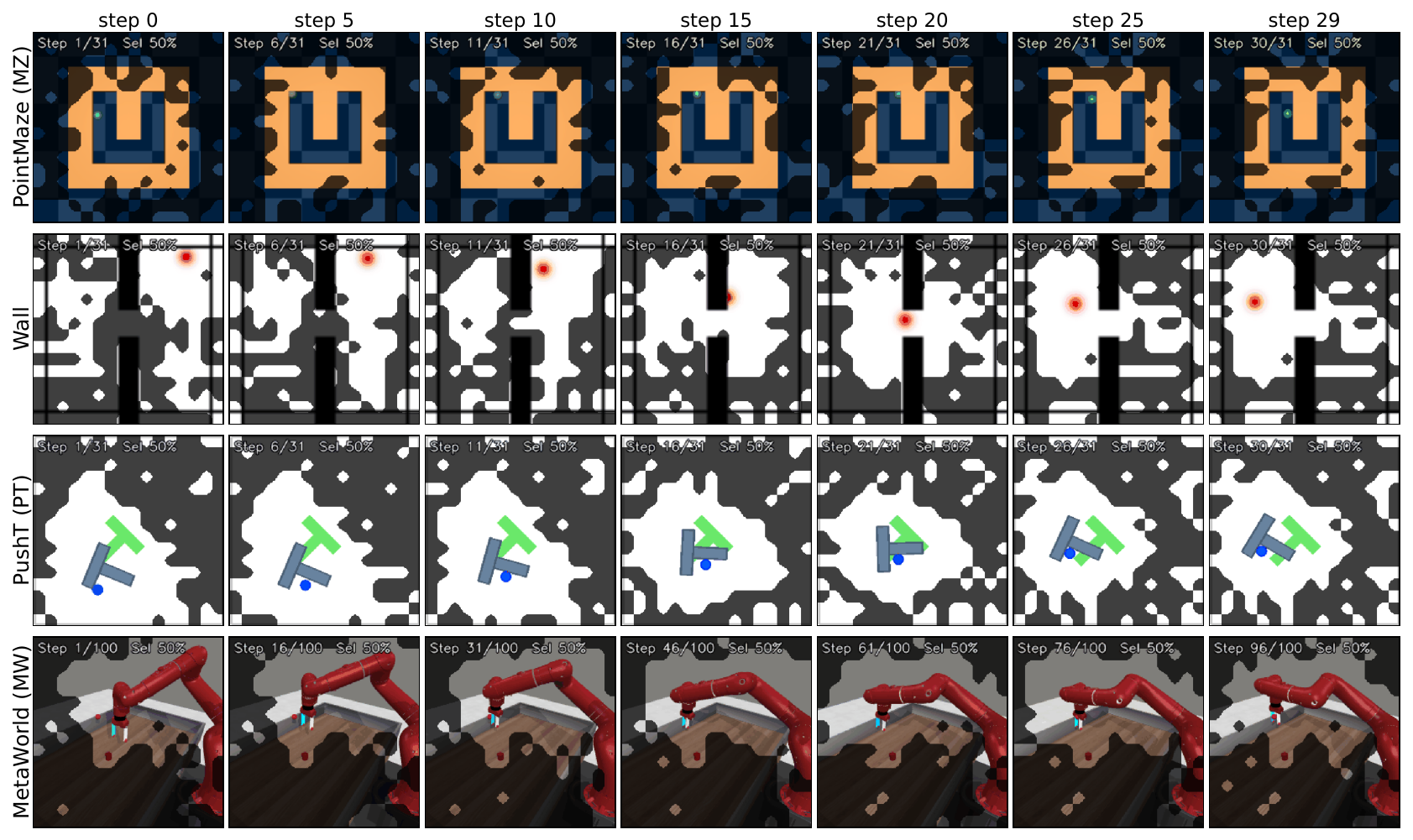}
\caption{\rev{\textbf{\costgrad{} selection across environments.} Per-row: selection overlay at seven timesteps through one episode (bright = kept, dim = dropped, $50\%$ kept throughout). \costgrad{} adapts to each environment's task structure: maze corridors on PointMaze, agent and wall on Wall, T-block and pusher on PushT, arm and object on MetaWorld. Decoded one-step predictions from the same sparse token sets are in Appendix~\ref{app:decoded_pred}.}}
\label{fig:cg_envs}
\end{figure}
\end{revision}

\begin{revision}
\pagebreak[0]
\begin{samepage}
\paragraph{Prediction difficulty is not task relevance.}
Prediction error selects visually hard-to-predict tokens, whereas planning needs tokens whose state affects the goal-reaching cost.
This distinction is clearest on PushT: the pusher and T-block are task-critical but move predictably under the action, so they receive low prediction error while still receiving high \costgrad{} score.
A quadrant analysis confirms that PushT uniquely contains many ``easy-but-critical'' tokens --- low prediction error but high planning gradient --- at $\sim\!2\times$ chance density, while the other three environments are depleted (Appendix~\ref{app:quadrant}).
Prediction-derived selection can therefore downweight precisely the objects that matter for control.
\par\end{samepage}
\end{revision}

\begin{revision}
\paragraph{Selection stability.}\label{sec:seed_stability}
Prediction-derived attention is unstable across equivalent predictors.
Three Wall AdaLN-Zero predictors trained with different seeds reach indistinguishable validation loss, but their top-$128$ attention selections overlap at only IoU $0.44 \pm 0.02$ and \predattn{} planning success varies by $50\pp$ across seeds.
\costgrad{} selections on the same predictors overlap at IoU $> 0.82$ and yield planning success within evaluation noise.
Planning-derived gradients therefore provide a more reliable selection signal than learned attention.
\end{revision}

\begin{revision}
\begin{samepage}
\paragraph{Probe-action robustness.}
CEM plans with action distributions far from $\bm{a}_0 = \bm{0}$, so a fixed probe action could in principle give unstable rankings.
\costgrad{} rankings remain stable across $10$ tested probe actions: $\rho = 0.918$ on AdaLN-Zero and $0.958$ on matched-concat (Appendix~\ref{app:nulls}).
On AdaLN-Zero, replacing the default zero-action probe with Gaussian-noise or CEM-mean probes changes Wall planning success by at most $\pm 5\pp$ (Appendix~\ref{app:probe}).
Stable token rankings do not guarantee reliable predictions after token removal; we next examine how sparse planning depends on the predictor's architecture (\S\ref{sec:swap}).
\par\end{samepage}
\end{revision}

\begin{samepage}
\subsection{Sparse planning depends on selector--architecture compatibility}
\label{sec:swap}

\rev{We evaluate selector--architecture compatibility on Wall and PointMaze, using three independently trained concat checkpoints per environment.}
\rev{The concat predictors share the encoder, training data, depth, heads, attention head dimension, MLP width, optimizer, and schedule with AdaLN-Zero.}
\rev{AdaLN supplies action-dependent scales, shifts, and gates at every block, with modulation parameters independent of the retained token set. Concat appends action features to each token at the input, before self-attention.}
\par\end{samepage}

\rev{Across both environments, \costgrad{} has no consistent advantage over Random on concat: the paired success-rate gaps are $-0.1\pm4.9\pp$ on Wall and $+2.0\pm4.5\pp$ on PointMaze (Appendix~\ref{app:nulls}, Table~\ref{tab:concat_replication}). This contrasts with the gains over Random on AdaLN-Zero (Table~\ref{tab:main}).}

\begin{table}[!t]
\ifrevisionmarks\color{red}\captionsetup{labelfont={color=red}}\fi
\centering\small
\caption{\rev{\textbf{Full and sparse planning on Wall.} AdaLN results are from Table~\ref{tab:main}; concat reports mean $\pm$ std across three training checkpoints, with $96$ episodes per method per checkpoint. CG$-$Full $> 0$: sparsification is profitable; $< 0$: net harmful.}}
\label{tab:swap}
\begin{tabular}{lccccc}
\toprule
Predictor & Full & \costgrad{} & Random & CG$-$Rand & CG$-$Full \\
\midrule
AdaLN-Zero        & $84.7 \pm 2.6$ & $92.0 \pm 1.8$ & $67.0 \pm 3.6$ & $+25.0$ & $+7.3$ \\
Concat (matched)  & $85.1 \pm 5.2$ & $80.0 \pm 2.9$ & $80.2 \pm 3.8$ & $-0.1$  & $-5.1$ \\
\bottomrule
\end{tabular}
\end{table}

\rev{Table~\ref{tab:swap} contrasts full and sparse planning on Wall. Full-token success is comparable between concat and AdaLN-Zero ($85.1\%$ vs.\ $84.7\%$). Under sparsification, \costgrad{} beats Random by $25.0\pp$ and Full by $7.3\pp$ on AdaLN-Zero, whereas it falls $5.1\pp$ below Full on concat. We next examine how token removal changes the action-dependent predictions used by CEM.}

\pagebreak[0]
\begin{samepage}
\begin{revision}
\paragraph{Action-pathway drift.}
CEM ranks candidate actions by their predicted outcomes.
Sparse Imagination~\citep{chun2025sparseimagination} identifies a related failure: fixed token subsets can create blind spots that make candidate actions indistinguishable.
We therefore examine whether token removal changes the spatial distribution of action effects across retained positions.
Let $f_i(\mathcal{U},\bm{a})$ denote the predictor's next-step visual output at position $i$ using token set $\mathcal{U}$ and action $\bm{a}$.
Holding context and proprioception fixed, define the action effect by
\begin{equation}
\bm{\delta}_i^{\mathcal{U}}=f_i(\mathcal{U},\bm{a})-f_i(\mathcal{U},\bm{0}).
\label{eq:action_effect_main}
\end{equation}
For the full set $\mathcal{P}=\{1,\ldots,P\}$ and kept set $\mathcal{S}$, normalize squared action effects over the same kept positions $i\in\mathcal{S}$:
\begin{equation}
p_i=\frac{\|\bm{\delta}_i^{\mathcal{S}}\|_2^2}{\sum_{j\in\mathcal{S}}\|\bm{\delta}_j^{\mathcal{S}}\|_2^2},
\qquad
q_i=\frac{\|\bm{\delta}_i^{\mathcal{P}}\|_2^2}{\sum_{j\in\mathcal{S}}\|\bm{\delta}_j^{\mathcal{P}}\|_2^2}.
\label{eq:action_distributions_main}
\end{equation}
We measure drift as $D_{\mathrm{KL}}(p\|q)=\sum_{i\in\mathcal{S}}p_i\log(p_i/q_i)$.
Using the same retained positions isolates redistribution from the absence of dropped outputs; normalization focuses on relative spatial magnitudes, discarding overall scale and vector direction.
Random selection at the same token budget provides a reference for generic sparsification.
The diagnostic does not itself bound planning error or establish that candidate-action rankings are preserved.
See Appendix~\ref{app:invariance} for details.
\end{revision}
\par\end{samepage}
\rev{Table~\ref{tab:kl_main} reports drift across three independently trained Wall checkpoints per architecture: \costgrad{} yields $(0.29\pm0.03)\times$ Random on AdaLN-Zero and $(1.33\pm0.17)\times$ on matched-concat, with the same direction in every checkpoint.}

\pagebreak[0]
\begin{samepage}
\paragraph{\rev{Coverage through random anchors.}}
\rev{To examine whether the concentration of selected tokens matters, we replace part of the gradient-selected subset with random anchors while keeping $K$ fixed.}
\rev{Mixing $20\%$ top-\costgrad{} tokens with $80\%$ random anchors ($\alpha{=}0.8$) brings concat drift to $(1.04\pm0.11)\times$ Random, while AdaLN remains below Random at $(0.44\pm0.02)\times$ (Table~\ref{tab:kl_main}).}
\rev{The smaller concat drift motivates broader coverage, but planning recovery is checkpoint-dependent (Appendix~\ref{app:nulls}); anchor mixing is therefore a diagnostic intervention, not a generally better selector.}
\par\end{samepage}

\begin{table}[!tb]
\ifrevisionmarks\color{red}\captionsetup{labelfont={color=red}}\fi
\centering\small
\caption{\rev{\textbf{Multi-seed action-pathway diagnostic on Wall.}
Mean $\pm$ sample standard deviation across three training seeds per architecture; $K{=}128$, $20$ observations $\times$ $5$ action probes per checkpoint. Selected/Random ratios below one indicate less drift than Random. $\alpha{=}0.8$ uses $80\%$ random anchors.}}
\label{tab:kl_main}
\begin{tabular}{llccc}
\toprule
Predictor & Selection & KL random & KL selected & Sel. / random \\
\midrule
AdaLN-Zero        & \costgrad{}    & $0.265\pm0.005$ & $0.077\pm0.009$ & $(0.29\pm0.03)\times$ \\
AdaLN-Zero        & $\alpha{=}0.8$ & $0.265\pm0.005$ & $0.116\pm0.009$ & $(0.44\pm0.02)\times$ \\
Concat (matched)  & \costgrad{}    & $0.287\pm0.017$ & $0.380\pm0.030$ & $(1.33\pm0.17)\times$ \\
Concat (matched)  & $\alpha{=}0.8$ & $0.287\pm0.017$ & $0.298\pm0.016$ & $(1.04\pm0.11)\times$ \\
\bottomrule
\end{tabular}
\end{table}

\pagebreak[0]
\begin{samepage}
\begin{revision}
\paragraph{Mechanistic interpretation.}
Our hypothesis is that AdaLN's repeated local conditioning reduces dependence of the action response on removed context, whereas concat relies on learned transformations of the action features supplied at the input; neither architecture guarantees this stability.
Reversing the ranking does not resolve the failure: keeping the bottom-gradient tokens collapses planning on both predictors (Appendix~\ref{app:nulls}).
Appendix~\ref{app:planning_preservation} separates visual cost restriction from changes caused by sparse dynamics.
For sparse planning, token relevance and the predictor's response to token removal must be evaluated together.
\par\end{revision}
\end{samepage}

\section{Related Work}
\label{sec:related}

\begin{revision}
\paragraph{World models and sparse planning.}
Token-based visual world models such as DINO-WM~\citep{zhou2025dinowm} and JEPA-WMs~\citep{terver2025jepawms} preserve a spatial grid of visual features for planning.
This spatial structure is useful for goal-conditioned control, but it makes CEM rollouts expensive because every candidate action sequence is evaluated over many tokens.
V-JEPA~2-AC~\citep{assran2025vjepa2} also uses an action-conditioned latent world model for image-goal robot planning.
Among compact-latent approaches, DreamerV3~\citep{hafner2023dreamerv3} learns a policy from imagined trajectories, while TD-MPC2~\citep{hansen2024tdmpc2} and LeWorldModel~\citep{maes2025lewm} optimize actions in lower-dimensional latent spaces.
\costgrad{} addresses the complementary setting: given a token-based world model, it reduces the number of spatial tokens used during planning without retraining the model.

Sparse Imagination~\citep{chun2025sparseimagination} combines randomized grouped attention during training with random token selection during planning.
It also analyzes spatial blind spots from importance-based selection that make candidate actions difficult to distinguish.
\costgrad{} instead uses planning-cost gradients to select a goal-conditioned subset in a fixed pretrained predictor and studies its interaction with action conditioning.
The approaches are complementary: sparse training improves tolerance to missing tokens, while planning-gradient selection identifies which tokens matter for the current goal.

\begin{samepage}
Contemporaneous Scope-WM~\citep{li2026scopewm} distills prediction-loss sensitivity into an action-conditioned selector and trains sparse dynamics with lightweight background updates.
Its selector--background-update interaction provides complementary evidence for matching selection to sparse dynamics.
\costgrad{} instead selects directly from the current goal's planning-cost gradients without retraining the predictor.
\par\end{samepage}

\paragraph{Token selection and importance estimation.}
Token-reduction methods for ViTs include token merging~\citep{bolya2023tome}, learned token pruning~\citep{rao2021dynamicvit}, class-token-driven reorganization~\citep{liang2022evit}, and patch dropout during training~\citep{liu2023patchdropout}.
These methods were primarily developed for single-pass perception, where the reduced token set supports one forward computation.
EfficientVLA~\citep{yang2025efficientvla} extends training-free token reduction to robot policies, combining task-aware attention scores with feature diversity.
Sparse visual planning is different: the selected tokens must support many action-conditioned rollouts inside CEM.

Gradient saliency has a long history, from input-gradient visualizations~\citep{simonyan2014deep} to axiomatic attribution methods such as Integrated Gradients~\citep{sundararajan2017axiomatic} and class-discriminative localization methods such as GradCAM~\citep{selvaraju2017gradcam}.
Sensitivity also guides computation reduction: Taylor-based pruning~\citep{molchanov2017pruning} approximates the loss change caused by removing convolutional filters, and ZOO-Prune~\citep{kim2026zooprune} estimates projected-feature sensitivity for training-free visual-token pruning.
\costgrad{} uses downstream planning-cost gradients to choose the token subset for repeated action-conditioned rollouts in a fixed predictor.
Its contribution couples this goal-conditioned selection rule with an analysis of whether the selected subset preserves the predictor's action response (\S\ref{sec:swap}).

\paragraph{Decision relevance and predictor compatibility.}
Objective mismatch~\citep{lambert2020objective}, value-aware model learning~\citep{farahmand2017value}, and value equivalence~\citep{grimm2020value} distinguish predictive fidelity from usefulness for downstream decisions.
\costgrad{} applies this principle to inference-time token selection in a fixed predictor: the planning objective determines which representation supports action search, without changing model training.

Underspecification~\citep{damour2022underspecification} describes how models with similar in-distribution metrics can differ under downstream use.
The attention-as-explanation debate~\citep{jain2019attention,wiegreffe2019attention} likewise highlights the need to evaluate attention's usefulness in context, rather than assume universal reliability or unreliability.
Our Wall seed study (\S\ref{sec:seed_stability}) demonstrates an operational consequence: predictors with matched validation loss yield unstable attention-based selections and planning outcomes, while planning-gradient selection is more stable in the same setting.

Feature-wise linear modulation (FiLM)~\citep{perez2018film} is a standard way to condition neural representations; adaptive normalization variants such as AdaLN-Zero~\citep{peebles2023dit} apply this idea inside transformer blocks.
DiT~\citep{peebles2023dit} compares adaptive normalization with cross-attention and in-context conditioning through appended tokens.
Our matched AdaLN-vs-channel-concat comparison (\S\ref{sec:swap}) adds stability under token removal as a criterion for action-conditioning design.
In our experiments, AdaLN supports pure planning-gradient sparsification, whereas concat makes concentrated subsets unstable and can require coverage-preserving selectors.
\par\end{revision}

\section{Discussion and Limitations}\label{sec:discussion}

\paragraph{When does sparse planning work?}
\rev{Our results identify two coupled design considerations for sparse planning: relevance to the control objective and compatibility with the predictor.
\costgrad{} derives token importance from the current goal, since importance for prediction need not coincide with importance for control.
In our matched comparison, \costgrad{} reduces measured action-effect drift relative to Random on AdaLN but increases it on concat, tracking its loss of planning advantage (\S\ref{sec:swap}).
Token-selection performance should therefore be assessed jointly with the predictor at the intended token budget.}

\paragraph{Limitations and open questions.}
\rev{Our evaluation covers four simulated environments, one encoder family (DINOv2 ViT-S/14) and one planner (CEM).
Physical-system evaluation with perception noise, calibration errors, and model mismatch remains future work.
Our selector--architecture finding is supported for the tested AdaLN and channel-wise concat predictors; its generality beyond this comparison remains untested.
Future work should evaluate FiLM, cross-attention, and prefix-style conditioning to test whether action-pathway drift predicts sparse-planning performance across conditioning mechanisms, and explore sparse training to improve robustness to gradient-selected subsets.
The strength of \costgrad{} is also environment-dependent: attention-based selection is competitive on PointMaze and PushT, while the gap is much larger on Wall and MetaWorld.
All experiments use fixed pretrained predictors and inference-time selection; online adaptation and learned selectors remain future work.
Selection uses a one-step probe, while CEM evaluates longer action sequences. Whether longer-horizon probes improve selection enough to justify their cost remains open.}

\paragraph{Design implication.}
A planner is part of a world model's operating distribution.
\rev{Token restriction can change the model's response to candidate actions, even when the retained tokens appear relevant to the task.
World models intended for sparse planning should be evaluated under the token subsets and action searches used at deployment, because full-token prediction quality alone does not establish their suitability for sparse control.}

\label{cr:last_main_page}
\clearpage
\bibliographystyle{plainnat}
\begingroup
\let\originalbibitem\bibitem
\renewcommand{\bibitem}[2][]{%
  \normalcolor
  \ifrevisionmarks
    \def\currentbibkey{#2}%
    \def\newbibkey{lambert2020objective}\ifx\currentbibkey\newbibkey\color{red}\fi
    \def\newbibkey{farahmand2017value}\ifx\currentbibkey\newbibkey\color{red}\fi
    \def\newbibkey{grimm2020value}\ifx\currentbibkey\newbibkey\color{red}\fi
    \def\newbibkey{yang2025efficientvla}\ifx\currentbibkey\newbibkey\color{red}\fi
    \def\newbibkey{assran2025vjepa2}\ifx\currentbibkey\newbibkey\color{red}\fi
    \def\newbibkey{molchanov2017pruning}\ifx\currentbibkey\newbibkey\color{red}\fi
    \def\newbibkey{kim2026zooprune}\ifx\currentbibkey\newbibkey\color{red}\fi
    \def\newbibkey{li2026scopewm}\ifx\currentbibkey\newbibkey\color{red}\fi
    \def\newbibkey{zagoruyko2017attentiontransfer}\ifx\currentbibkey\newbibkey\color{red}\fi
  \fi
  \originalbibitem[#1]{#2}%
}
\bibliography{references}
\endgroup

\clearpage

\appendix

\section{Environments}
\label{app:envs}

We evaluate on four continuous-control benchmarks following \citet{terver2025jepawms}.
All four are visual planning tasks where the agent is given an initial frame and a goal frame and must produce an action sequence that drives the environment to the goal.

\paragraph{PointMaze (MZ).}
A 2D point navigates \rev{a U-Maze configuration} drawn from \citet{terver2025jepawms}'s suite (the underlying maze layout follows the D4RL maze suite, \citealp{fu2020d4rl}).
Action space: 2-dimensional continuous (x, y \rev{motor controls}).
Visual: a $224\times 224$ rendering of the agent and maze walls.
Episode horizon: 30 environment steps; success = reaching within $0.5$ of the goal location.

\paragraph{Wall.}
A 2D agent navigates around a vertical wall with a single horizontal gap to reach a goal on the other side.
Action space: 2-dimensional continuous; frame-skip $5$, so each predictor step receives a $10$-dim concatenated action.
Visual: $224\times 224$ rendering with the agent as a small red dot, the wall as a black bar, and the gap as the wall's discontinuity.
Episode horizon: 30; success = $L_2$ distance to goal \rev{$< 4.5$}.

\paragraph{PushT (PT).}
A 2D pusher pushes a T-shaped block to a target pose, following the PushT environment used by \citet{chi2023diffusionpolicy}.
Action space: 2-dimensional continuous (pusher target position).
Visual: $224\times 224$ rendering with a small blue pusher dot, a blue T-block, and a green target T over a plain light-gray background.
Episode horizon: 30; success = \rev{$L_2$ error in the concatenated pusher/block positions $< 20$ and block-angle error $< \pi/9$}.

\paragraph{MetaWorld (MW).}
A 3D 7-DoF arm performs \rev{the Reach task from} MetaWorld's training suite~\citep{yu2020metaworld}.
Action space: \rev{4-dimensional continuous (three Cartesian end-effector increments and one gripper command)}.
Visual: $224\times 224$ third-person view; proprio: \rev{4-dimensional vector (end-effector position and gripper opening)}.
Episode horizon: \rev{100 environment steps}; success per the MetaWorld task definition (\rev{end-effector--target distance $\leq 0.05$}).

\section{Implementation and hyperparameters}
\label{app:hyperparams}

\paragraph{Encoder.}
Frozen DINOv2 ViT-S/14~\citep{oquab2024dinov2}; we do not fine-tune any encoder weights.
Input image $224 \times 224$; patch size $14$; produces a $16 \times 16 = 256$ grid of $384$-dimensional tokens per frame.

\paragraph{Predictor.}
Six-block transformer with $400$-dimensional internal embedding and $16$ heads.
RoPE positional encoding; GELU activation in the FFN, SiLU in the AdaLN modulation MLP; per-block AdaLN modulation parameters $(\bm{\gamma}_1, \bm{\beta}_1, \bm{g}_1, \bm{\gamma}_2, \bm{\beta}_2, \bm{g}_2)$ derived from the action via a single linear layer.
Action embedding dimension $400$ (matching internal embedding); action tokens: $1$.
Total predictor parameters: $\approx 17.6$M.

\paragraph{Concatenation predictor (matched-concat).}
Same depth ($6$ blocks), heads ($16$), attention head dimension ($25$), MLP hidden dimension ($1600$), GELU activation, RoPE positional encoding, training data, and optimizer as the AdaLN predictor.
Action conditioning is replaced: a $10$-dimensional frame-skip action vector is channel-wise concatenated to each visual token before the q/k/v projections (DINO-WM-style).
The only architectural difference is the action pathway.

\paragraph{Planner.}
Cross-Entropy Method (CEM, \citealp{rubinstein1999cem}): horizon $H{=}6$, $300$ candidate sequences per iteration, $30$ iterations, $10$ elites for refit.
Cost: L$_2$ between predicted final-step encoding and goal encoding (cosine and L$_1$ alternatives in Appendix~\ref{app:cost_function}).
$\alpha = 0.1$ proprio weight when applicable.

\paragraph{Pretraining.}
Our retrained predictors are trained for $20$ epochs with \rev{AdamW}, learning rate \rev{$5\times 10^{-4}$}, and batch size \rev{$8$ per GPU}, following the reference JEPA-WMs hyperparameters.
Unless otherwise stated, main results use our retrained checkpoints; \costgrad{} itself is training-free and can also be applied directly to released checkpoints.

\begin{revision}
\paragraph{Objective and normalization.}
For retained positions $\mathcal S$, $K=|\mathcal S|$, the default final-step CEM cost is
\[
J_{\mathcal S}(A)=\frac{1}{KD}\sum_{j\in\mathcal S}\|\hat{\bm z}_{H,j}(A)-\bm z_{g,j}\|_2^2
+\frac{\alpha}{d_p}\|\hat{\bm p}_H(A)-\bm p_g\|_2^2,\qquad \alpha=0.1,
\]
where $\bm p$ is the flattened proprioceptive representation and $d_p$ its number of entries; the second term is omitted when proprioception is absent.
Full uses all $P$ positions. Both terms average over their own feature entries, so reducing $K$ does not implicitly reduce the visual term's weight.
The selection probe uses only the full-grid visual sum of squared errors; replacing this sum by a mean would uniformly rescale its gradients without changing their ranking.
RGB inputs are scaled to $[0,1]$ and standardized with ImageNet channel statistics before DINOv2 encoding; its output features receive no additional feature normalization.
Proprioceptive inputs are standardized using dataset means and standard deviations before encoding, and actions use the training-normalized coordinates.
Probe-based selectors use zero action and proprioceptive inputs in these coordinates; CEM rollout uses the observed proprioceptive input.

\paragraph{Attention aggregation.}
Let $A^{(h)}$ be the last block's attention matrix for head $h$, after the scaled query--key softmax with block-causal masking, and let $k(t,j)$ index spatial position $j$ in context frame $t$.
With $M$ heads and $T$ context frames, \predattn{} uses
\[
s_j^{\mathrm{attn}}=\frac{1}{MT}\sum_{h=1}^{M}\sum_{t=1}^{T}\sum_q A^{(h)}_{q,k(t,j)}.
\]
The query sum includes all queries; only spatial keys are ranked. Heads and frames receive equal weight, with no entropy weighting.

\paragraph{Prediction-gradient target.}
For an observed context with $T\geq2$, \predgrad{} computes
\[
\mathcal L_{\mathrm{pred}}=\|f(\bm z_{1:T-1},\bm 0)_{\mathrm{last}}-\operatorname{stopgrad}(\bm z_T)\|_2^2,
\qquad s_j^{\mathrm{pred}}=\|\nabla_{\bm z_{T-1,j}}\mathcal L_{\mathrm{pred}}\|_2.
\]
Here $f$ returns visual predictions and the subscript selects its last temporal output.
The target is a detached encoding from observed history, not an outcome obtained by executing the probe action; the historical action that produced it is not substituted for the zero probe.
Top-$K$ spatial indices from this retrospective residual are applied to the current observation and goal.

\paragraph{Token-merging adaptation.}
ToMe partitions the current grid into checkerboard source and destination sets, matches each source to its most cosine-similar destination, and merges the $P-K$ highest-similarity source--destination pairs.
Each group is replaced by the equal-weight average of its original features at the destination's original positional index.
The mapping is computed once per planning step and applied to every observed context frame; subsequent rollout predicts only the surviving positions without rematching.
The visual cost compares these predictions with the unmerged goal features at the same positions, averaging uniformly over surviving tokens without merge-size weights.
\end{revision}

\section{Decoded one-step predictions from \costgrad{}-50\% sparse tokens}
\label{app:decoded_pred}

\begin{figure}[!tb]
\centering
\includegraphics[width=\linewidth]{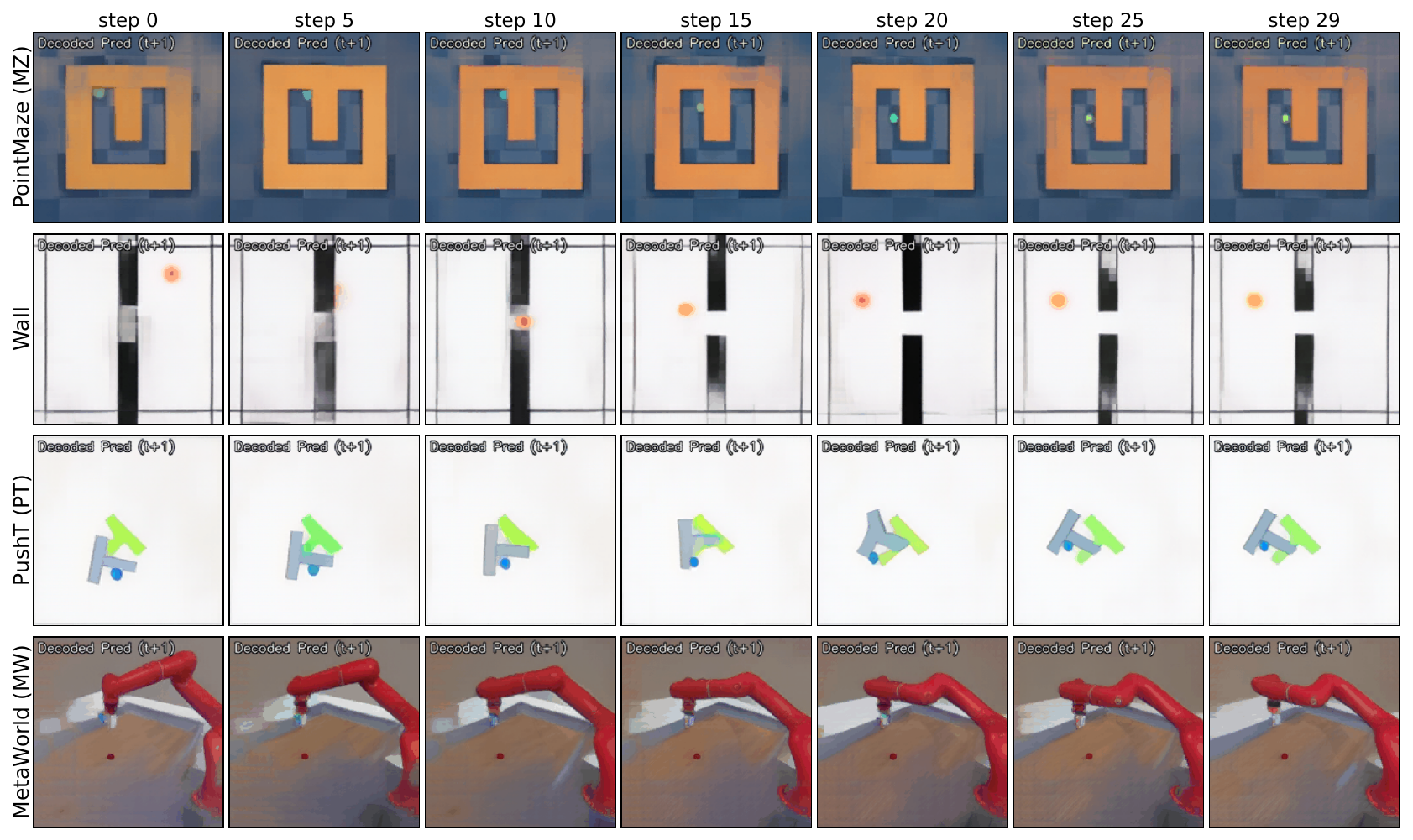}
\caption{\textbf{Decoded one-step predictions from \costgrad{}-50\% sparse tokens.} Per-row: the predictor's decoded $t{+}1$ output at seven timesteps through one episode, when only the top-$128$ \costgrad{}-scored tokens are kept (corresponding to the selection overlays in Figure~\ref{fig:cg_envs}).}
\label{fig:cg_envs_decoded}
\end{figure}

Figure~\ref{fig:cg_envs_decoded} shows the predictor's decoded next-step predictions from the \costgrad{}-selected $50\%$ token subsets visualized in Figure~\ref{fig:cg_envs}.
On all four environments, the prediction quality at $50\%$ tokens is qualitatively faithful: agent and goal positions are recovered on PointMaze and Wall, the T-block and pusher poses are recovered on PushT, and the arm/object configuration is recovered on MetaWorld.
Visual artifacts from missing tokens are concentrated in background regions \costgrad{} chose to drop.

\FloatBarrier
\begin{samepage}
\section{\rev{Action conditioning and token-removal sensitivity}}
\label{app:invariance}

\rev{We summarize the conditioning mechanisms and define the drift diagnostic used in \S\ref{sec:swap}. Let $\bm{z}_i$ denote a block's input token, $\bm{a}$ the action input, and $\mathcal{S}$ the retained spatial indices. The schematic block updates below omit temporal indexing and proprioceptive channels.}

\textbf{AdaLN-Zero block} \rev{(predictor used in \S\ref{sec:setup})}:
\begin{align}
\bm{h}_i &= \bm{z}_i + \bm{g}_1(\bm{a}) \odot \mathrm{Attn}_i\big(\{(1 + \bm{\gamma}_1(\bm{a})) \odot \mathrm{LN}(\bm{z}_j) + \bm{\beta}_1(\bm{a}) : j \in \mathcal{S}\}\big), \\
T_i^{(\ell+1), \text{AdaLN-Zero}} &= \bm{h}_i + \bm{g}_2(\bm{a}) \odot \mathrm{MLP}\big((1 + \bm{\gamma}_2(\bm{a})) \odot \mathrm{LN}(\bm{h}_i) + \bm{\beta}_2(\bm{a})\big).
\end{align}
\par\end{samepage}
The action enters at six per-token modulation parameters $(\bm{\gamma}_1, \bm{\beta}_1, \bm{g}_1, \bm{\gamma}_2, \bm{\beta}_2, \bm{g}_2)$, all functions of $\bm{a}$ alone, broadcast per-token.
\begin{revision}
The gates $\bm{g}_1,\bm{g}_2$ are zero-initialized, so each block starts as the identity (the ``Zero'' in AdaLN-Zero).
The action-dependent modulation is broadcast across tokens; this structure, rather than the zero initialization, is what \S\ref{sec:swap} returns to when comparing AdaLN against concat conditioning.
\end{revision}

\textbf{Concat block} (DINO-WM style; channel-wise feature concatenation): with $\bm{z}_i^{\mathrm{aug}} := [\bm{z}_i; \pi(\bm{a})] \in \reals^{D + D_a}$,
\begin{align}
\bm{h}_i &= \bm{z}_i^{\mathrm{aug}} + \mathrm{Attn}_i\big(\{\mathrm{LN}(\bm{z}_j^{\mathrm{aug}}): j \in \mathcal{S}\}\big), \\
T_i^{(\ell+1), \text{concat}} &= \bm{h}_i + \mathrm{MLP}(\mathrm{LN}(\bm{h}_i)).
\end{align}
The action enters once, channel-wise per token, before the q/k/v projections.

\begin{revision}
\paragraph{Dependence on the retained tokens.}
AdaLN's modulation parameters depend on the action alone, but its modulated tokens still enter self-attention. In both architectures, removing tokens changes attention normalization and aggregated values, so action effects at the block output can change. The conditioning mechanisms therefore motivate an empirical comparison of sensitivity to token removal; subset-independent modulation parameters alone do not establish output invariance.

\paragraph{Output-level action-pathway drift.}
Let $f_i(\mathcal{U},\bm{a})$ be the predictor's next-step visual output at position $i$ using token set $\mathcal{U}$. With context and proprioception fixed, define
\begin{equation}
\bm{\delta}_i^{\mathcal{U}} = f_i(\mathcal{U},\bm{a})-f_i(\mathcal{U},\bm{0}).
\end{equation}
For $\mathcal{P}=\{1,\ldots,P\}$ and $i\in\mathcal{S}$, normalize the squared action effects of both predictors over the same retained positions:
\begin{equation}
p_i=\frac{\|\bm{\delta}_i^{\mathcal{S}}\|_2^2}{\sum_{j\in\mathcal{S}}\|\bm{\delta}_j^{\mathcal{S}}\|_2^2},
\qquad
q_i=\frac{\|\bm{\delta}_i^{\mathcal{P}}\|_2^2}{\sum_{j\in\mathcal{S}}\|\bm{\delta}_j^{\mathcal{P}}\|_2^2}.
\end{equation}
We report $D_{\mathrm{KL}}(p\|q)=\sum_{i\in\mathcal{S}}p_i\log(p_i/q_i)$, with $10^{-10}$ numerical stabilization in the implementation. This measures changes in the spatial distribution of squared action-effect magnitudes; it does not measure their vector directions or a common change in magnitude.

\paragraph{Diagnostic design.}
Subtracting the zero-action output removes action-independent offsets, so the comparison concerns the response to an action rather than raw prediction agreement.
We compare finite action responses of the same predictor under full and sparse inputs.
Summing squared responses over feature channels produces a nonnegative spatial map, analogous to the squared-activation maps used in attention transfer~\citep{zagoruyko2017attentiontransfer}.
We normalize over the same retained positions to compare relative spatial allocation without counting missing positions or a uniform rescaling as drift.
Our sum normalization and KL divergence differ from attention transfer's normalized-map distance: $D_{\mathrm{KL}}(p\|q)$ penalizes sparse-response mass where the full predictor assigns little relative mass.
This is an output-response diagnostic, not a direct measurement of internal computational pathways or a guarantee of action-ranking preservation.

\paragraph{Empirical architecture interaction.}
Across three Wall training seeds per architecture, Table~\ref{tab:kl_main} reports a \costgrad{}-selected/Random drift ratio of $0.29\pm0.03$ for AdaLN-Zero and $1.33\pm0.17$ for matched-concat, with the direction holding for every checkpoint. The proposed conditioning-based explanation is discussed in \S\ref{sec:swap}.
\end{revision}

\section{Probe action sensitivity (\costgrad{} and \predattn{})}
\label{app:probe}

Both \costgrad{} and \predattn{} require a probe action sequence to compute their importance scores.
\rev{Both use a zero-action probe by default, corresponding to the mean action in normalized training space and providing a neutral starting point before action search. For \costgrad{}, replacing zero with small Gaussian noise changes planning success by at most $5\pp$ across the four environments. A CEM-mean probe was additionally tested on Wall, also within $5\pp$; we have not tested CEM-mean on every environment.}
The probe-stability of \costgrad{}'s selection extends to the per-token level: rank Spearman across $10$ probe actions is $\rho > 0.9$ on both AdaLN-Zero and the matched-concat predictor (Wall, seed 1; \rev{Appendix~\ref{app:nulls}}).

\section{Cost function robustness}
\label{app:cost_function}

\costgrad{}'s benefit is robust across the choice of planning cost function; in fact the benefit is largest when the cost is poorest.
Table~\ref{tab:cost} reports Wall planning success with three cost functions: L$_2$ (default), cosine similarity, and L$_1$.
\rev{This ablation varies the CEM planning cost while retaining the L$_2$ selection probe for \costgrad{} in every row.}

\begin{table}[!tb]
\centering\small
\caption{Wall planning success vs.\ choice of CEM cost function.
\rev{The selector uses an L$_2$ probe throughout; its largest gain is under the L$_1$ CEM cost, where full-token planning performs worst.}}
\label{tab:cost}
\begin{tabular}{lccc}
\toprule
Cost & Full & \costgrad{} ($50\%$) & $\Delta$ \\
\midrule
L$_2$    & $84.7$ & $92.0$ & $+7.3$ \\
Cosine   & $82.3$ & $87.5$ & $+5.2$ \\
L$_1$    & $47.9$ & $70.8$ & $+22.9$ \\
\bottomrule
\end{tabular}
\end{table}

\begin{revision}
\paragraph{Cross-cost selection IoU.}
In a separate diagnostic, we vary the selection-probe objective itself: top-$128$ Jaccard IoU between L$_2$ and cosine selections is $0.917$; between L$_2$ and L$_1$ it is $0.841$.
The three probe objectives therefore select substantially overlapping token sets in this diagnostic.
\end{revision}

\FloatBarrier
\section{Matched compute baseline}
\label{app:matched_compute}

A natural concern: \costgrad{} saves predictor compute by using fewer tokens, but the saved compute could alternatively be spent on more CEM iterations or candidates.
We test this directly on Wall: \costgrad{} at $50\%$ tokens with half the CEM iterations ($15$ instead of $30$) achieves $92.0 \pm 5.1\%$ planning success across $3$ eval seeds (per-seed: $89.58, 88.54, 97.92$) at $\sim\!5\times$ overall wall-clock speedup ($2.6\times$ from token sparsity $\times$ $2\times$ from halving CEM iterations; Appendix~\ref{app:timing}).
The mean matches the standard \costgrad{}-50\% rate ($92.0 \pm 1.8\%$, Table~\ref{tab:main}) and exceeds the $84.7\%$ Full baseline by $+7.3\pp$.
Eval-seed variance is wider here than for standard \costgrad{}-50\% ($\sigma = 5.1$ vs.\ $\sigma = 1.8$), but the worst-case seed ($88.54\%$) still beats Full.

\paragraph{Full at half CEM iters as a matched-compute reference.}
The natural reviewer comparison is whether halving \emph{Full}'s CEM iterations would achieve a similar speedup--success trade-off without any token selection.
Three eval seeds at Full ($K{=}256$) with $15$ iterations on Wall give $84.38, 83.33, 85.42 = 84.4 \pm 1.0\%$, within eval-seed noise of the standard $84.7 \pm 2.6\%$ for $2\times$ wall-clock speedup ($88.6 \to 44.3$~s; predictor compute scales linearly with iter count and \costgrad{} adds no overhead in this case since it is not used).
\costgrad{}-50\%-with-half-iters beats Full-with-half-iters by $+7.6\pp$ ($92.0$ vs.\ $84.4$) at the same iteration budget and a further $2.6\times$ per-iter cost reduction.
Token selection and increased search are therefore complementary: they address different sources of inefficiency.

\section{\costgrad{} operating curve}
\label{app:cg_curve}

\begin{figure}[!tb]
\centering
\includegraphics[width=0.7\linewidth]{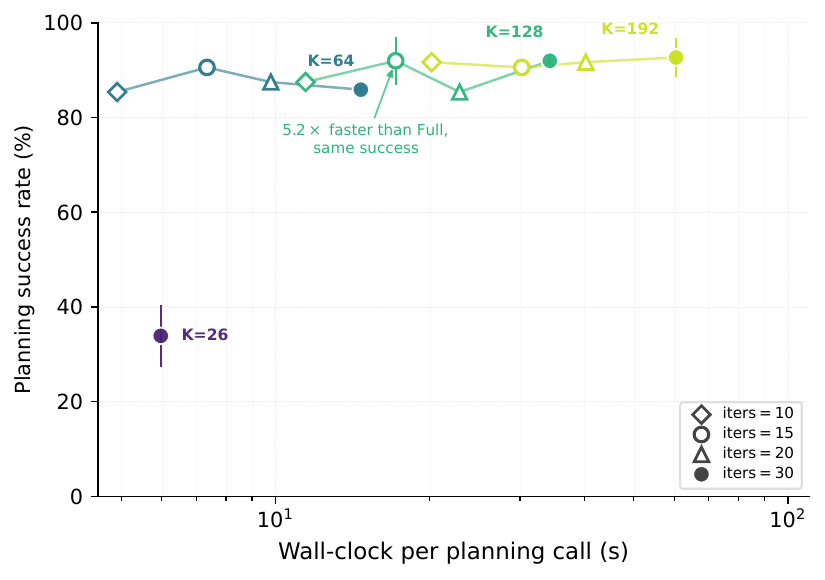}
\caption{\textbf{\costgrad{} operating curve on Wall AdaLN-Zero.}
Every measured $(K, \text{iters})$ \costgrad{} cell. Color encodes $K$ cluster (sequential viridis: purple $K{=}26$, teal $K{=}64$, green $K{=}128$, yellow $K{=}192$); marker shape encodes iter count ($\Diamond{=}10$, open $\bigcirc{=}15$, $\bigtriangleup{=}20$, filled $\bullet{=}30$). Thin lines connect points within each $K$ cluster (sorted by iter count). $K$ labels at the iters$=30$ marker of each cluster. The headline operating point ($K{=}128$ with iters$=15$, $\sim$17~s, $92.0\pm 5.1\%$) gives a $5.2\times$ wall-clock speedup over Full at higher mean success.}
\label{fig:cg_curve}
\end{figure}

Figure~\ref{fig:cg_curve} shows every measured \costgrad{} operating point on AdaLN-Zero, sweeping both token budget $K \in \{26, 64, 128, 192\}$ and CEM iteration count iters $\in \{10, 15, 20, 30\}$.
Color encodes the token budget $K$ cluster; marker shape encodes iter count ($\Diamond{=}10$, open $\bigcirc{=}15$, $\bigtriangleup{=}20$, filled $\bullet{=}30$).
Thin lines connect points within each $K$ cluster, in increasing iter order.

Two qualitative features support the practical claim that \costgrad{} ``saturates'' at the standard $50\%$ operating point.
First, planning success is roughly flat across $K \in \{64, 128, 192\}$ at iters$=30$ ($85.9\%$, $92.0\%$, $92.7\%$), so the marginal benefit of doubling tokens beyond $K{=}64$ is small.
Second, within each $K$ cluster, halving iterations ($30 \to 15$) does not consistently reduce success --- measurements at iters$\in\{10, 15, 20\}$ scatter around the iters$=30$ mean rather than tracking it monotonically.
Reducing CEM iterations therefore acts mostly as a horizontal slide along the time axis at roughly fixed success, until iterations get very low.

\suppressfloats[t]
\begin{revision}
\begin{samepage}
\section{\rev{Additional diagnostic checks}}
\label{app:nulls}

Sparse planning retains $K{=}128$ of $256$ tokens. Table~\ref{tab:concat_replication} gives the per-training-seed concat results on Wall and PointMaze summarized in \S\ref{sec:swap}; the remaining checks use Wall. Appendix~\ref{app:invariance} defines the action-pathway diagnostic, and Table~\ref{tab:kl_main} reports its multi-seed results.
\par\end{samepage}

\begin{table}[t]
\ifrevisionmarks\color{red}\captionsetup{labelfont={color=red}}\fi
\centering\small
\caption{\rev{\textbf{Concat planning across environments and training seeds.} Paired \costgrad{}--Random success-rate gaps (pp), with $K{=}128$ of $256$ tokens and $96$ episodes per method per checkpoint. Mean $\pm$ std is across the three training seeds.}}
\label{tab:concat_replication}
\begin{tabular}{lrrrr}
\toprule
Environment & Seed 1 & Seed 2 & Seed 3 & Mean $\pm$ std \\
\midrule
Wall      & $+2.1$ & $+3.2$ & $-5.7$ & $-0.1\pm4.9$ \\
PointMaze & $+4.1$ & $-3.2$ & $+5.2$ & $+2.0\pm4.5$ \\
\bottomrule
\end{tabular}
\end{table}

\pagebreak[0]
\begin{samepage}
\paragraph{Probe-action stability.}
Across $30$ observations and $45$ pairs of $10$ probe actions per observation, per-token rank Spearman is $0.918 \pm 0.058$ on AdaLN-Zero and $0.958 \pm 0.039$ on matched-concat (one checkpoint per architecture; mean $\pm$ standard deviation over all $1{,}350$ pairwise correlations).
The probes comprise zero, four positive unit-axis directions, and five Gaussian samples with $\sigma{=}0.5$.
Both predictors maintain stable rankings under these probes despite their different responses to token removal.
\par\end{samepage}

\pagebreak[0]
\begin{samepage}
\paragraph{Inverted-\costgrad{} sanity check.}
Keeping the bottom-$K$ tokens by gradient norm yields $0/60$ successful episodes on AdaLN-Zero (the run was stopped after $60$ of $96$ planned episodes) and $1/96$ ($1.04\%$) on matched-concat.
Reversing the ranking therefore collapses planning on both predictors; it does not recover \costgrad{}'s advantage over Random on concat.
\par\end{samepage}

\pagebreak[0]
\begin{samepage}
\paragraph{Anchor mixing.}
The selector keeps $(1{-}\alpha)K$ top-\costgrad{} tokens and fills the remaining $\alpha K$ slots uniformly at random from the unselected tokens.
At $\alpha{=}0.8$, planning improves over Random on two of three matched-concat training checkpoints.
On the first checkpoint, success is $94.1\pm2.4\%$ versus Random's $79.5\pm2.1\%$ (three evaluation seeds).
On the other two checkpoints, the respective anchor/Random results are $89.6/76.0\%$ and $76.0/83.3\%$ (one evaluation seed each); all evaluations use $96$ episodes per seed.
Together with Table~\ref{tab:kl_main}, these results show that restoring drift toward the Random baseline does not guarantee a planning gain.
We therefore use anchor mixing as a diagnostic of selector--architecture compatibility; its planning benefit is checkpoint-dependent.
\par\end{samepage}
\end{revision}

\section{Quadrant analysis methodology}
\label{app:quadrant}

For each environment we compute, on $30$ observations, the per-token prediction error $e_j = \|\mathrm{pred}_j - \mathrm{target}_j\|^2$ and the \costgrad{} score $s_j$.
\begin{revision}
For each observation, the easy-but-critical intersection contains tokens in the bottom $25\%$ of prediction error and the top $25\%$ of \costgrad{} score.
The reverse intersection contains tokens in the top $25\%$ of prediction error and the bottom $25\%$ of \costgrad{} score; these two intersections do not partition all $256$ tokens.
Under independence, the expected count in each intersection is $0.25 \times 0.25 \times 256 = 16$.
The easy-but-critical intersection measures disagreement between prediction error and planning sensitivity: tokens with low prediction error but high planning-cost gradient.
\end{revision}

\begin{table}[!tb]
\centering\small
\caption{\rev{Easy-but-critical token counts per environment ($30$ obs each, $P{=}256$): bottom-quartile prediction error intersected with top-quartile \costgrad{} score. Counts are mean $\pm$ std; the independence reference is $16$. PushT is the only environment with enrichment above this reference; the other three are depleted.}}
\label{tab:quadrant}
\begin{tabular}{lcc}
\toprule
Environment & Easy + critical (count) & Multiple of expected \\
\midrule
\textbf{PushT (PT)}    & $\mathbf{31.9 \pm 4.3}$ & $\mathbf{1.99\times}$ \\
PointMaze (MZ) & $10.7 \pm 2.8$ & $0.67\times$ \\
Wall            & $8.5 \pm 4.3$  & $0.53\times$ \\
MetaWorld (MW)  & $3.2 \pm 2.2$  & $0.20\times$ \\
\bottomrule
\end{tabular}
\end{table}

\begin{revision}
PushT has almost twice the independence-reference count of easy-but-critical tokens; Wall, MZ, and MW have fewer than expected.
This supports the interpretation that predictable motion can still matter for control: on PushT, the pusher and T-block can have low prediction error while receiving high planning-gradient scores.
Depletion in the other environments describes this particular intersection, rather than establishing agreement between the complete token rankings.
The reverse intersection is below the independence reference in every environment ($0.91\times$ on PushT; $0.20$--$0.87\times$ elsewhere), so it does not show the same enrichment.
\end{revision}

\section{Timing breakdown}
\label{app:timing}

We measure per-component latency on the AdaLN-Zero predictor \rev{using the Wall checkpoint and a} single \rev{H100} GPU, mean $\pm$ std over 5 warm-cache trials with 3 warmups discarded.

\begin{table}[!tb]
\centering\small
\caption{Timing breakdown for one CEM planning step ($300$ candidates $\times$ $30$ iterations $\times$ $H=6$). \costgrad{}'s extra forward+backward is a one-time cost amortized over $30$ CEM iterations, while the per-iteration cost drops by $2.61\times$ from the predictor-compute reduction. The reported total includes predictor rollout time and \costgrad{} overhead; the shared $7.64$~ms encoder cost is shown separately and is negligible at this scale. Net wall-clock speedup at $K=128$ vs.\ $K=256$ is $2.60\times$.}
\label{tab:timing}
\begin{tabular}{lcc}
\toprule
Component                                                                  & K=256 (Full) & K=128 (\costgrad{}) \\
\midrule
Encode context + goal (ms)                                                 & $7.64$       & $7.64$ \\
\costgrad{} forward + backward (ms)                                        & ---          & $27.53$ \\
Predictor forward at batch $300$ (ms)                                      & $492.25$     & $188.90$ \\
\midrule
Per CEM iter ($\times H{=}6$, ms)                                          & $2953.51$    & $1133.42$ \\
Total predictor over CEM step ($\times 30$ iters, s)                       & $88.61$      & $34.00$ \\
\midrule
\textbf{Planning call total, excl.\ shared encoder (s)}                                      & $\mathbf{88.61}$ & $\mathbf{34.04}$ \\
\bottomrule
\end{tabular}
\end{table}

The $2.60\times$ wall-clock speedup exceeds the naive $2\times$ that linear scaling in token count would predict.
The reason is that predictor compute mixes attention ($O(N^2)$ in token count) and MLPs ($O(N)$): at $N=128$ versus $N=256$, attention flops drop by $4\times$ while MLP flops drop by $2\times$, and the weighted combination on this architecture lands at $\sim\!2.6\times$.
The \costgrad{} forward+backward overhead ($27.53$ ms) amortizes to $\sim\!1$ ms per CEM iteration over $30$ iterations, an order of magnitude smaller than the per-iteration savings ($\sim 1820$ ms), so the overhead is operationally invisible.

\begin{revision}
\section{Public-checkpoint control}
\label{app:dropin}

We evaluate \costgrad{} on released JEPA-WMs checkpoints~\citep{terver2025jepawms} without retraining (Table~\ref{tab:dropin}).

\begin{table}[!b]
\ifrevisionmarks\color{red}\captionsetup{labelfont={color=red}}\fi
\centering\small
\caption{\rev{Planning success (\%) on released checkpoints without retraining. Mean $\pm$ standard deviation over three evaluation seeds, with $96$ episodes per seed. Full uses all $256$ tokens; \costgrad{} and Random keep $128$ ($50\%$).}}
\label{tab:dropin}
\begin{tabular}{lccc}
\toprule
Environment & Full & \costgrad{} ($50\%$) & Random ($50\%$) \\
\midrule
PointMaze & $87.7 \pm 7.7$ & $86.3 \pm 11.9$ & $70.7 \pm 15.1$ \\
Wall      & $87.0 \pm 2.8$ & $87.0 \pm 2.2$  & $75.7 \pm 4.6$ \\
PushT     & $66.3 \pm 2.6$ & $67.3 \pm 3.4$  & $21.0 \pm 9.0$ \\
MetaWorld (Reach) & $58.3 \pm 10.0$ & $63.2 \pm 5.2$ & $20.8 \pm 9.0$ \\
\bottomrule
\end{tabular}
\end{table}

\end{revision}

\clearpage
\begin{revision}
\section{An analytical perspective on token selection}
\label{app:planning_preservation}

The arguments below concern the visual squared-distance cost; the evaluated planner also includes weighted proprioceptive error.

\paragraph{Local relevance.}
At a fixed probe, write $f=f(\bm z,\bm a_0)$ and $G_j=\partial f/\partial\bm z_j$.
With fixed prediction target $\bm z_{\mathrm{next}}$ and goal $\bm z_g$, squared losses give
\begin{equation}
\nabla_{\bm z_j}L_{\mathrm{plan}}=2G_j^\top(f-\bm z_g),
\qquad
\nabla_{\bm z_j}L_{\mathrm{pred}}=2G_j^\top(f-\bm z_{\mathrm{next}}).
\label{eq:residual_alignment}
\end{equation}
Exact prediction makes the second gradient zero without requiring the first to vanish.
For the fixed full-token computation, let $g_j=\nabla_{\bm z_j}L_{\mathrm{plan}}$ and let $\mathcal S$ contain the $K$ retained tokens.
Perturb only discarded tokens, independently within $\|\Delta\bm z_j\|_2\leq\eta$.
Then
\begin{equation}
\max_{\{\|\Delta\bm z_j\|_2\leq\eta:\,j\notin\mathcal S\}}
\left|\sum_{j\notin\mathcal S}g_j^\top\Delta\bm z_j\right|
=\eta\sum_{j\notin\mathcal S}\|g_j\|_2.
\label{eq:local_selection_bound}
\end{equation}
Cauchy--Schwarz gives the upper bound; aligned perturbations attain it.
Keeping the $K$ largest norms minimizes this linearized criterion.
Physical token removal also changes attention and subsequent rollouts, which this local perturbation model does not represent.

\paragraph{Exact cost decomposition.}
Fix an observation, goal, horizon $H$, and token set $\mathcal S$ shared across action candidates.
Let $\Phi_{\mathcal P}(A)$ and $\Phi_{\mathcal S}(A)$ denote the full and sparse terminal predictions for action sequence $A$, with $\mathcal P=\{1,\ldots,P\}$.
For this visual component, sums and means induce the same ranking because the token count is fixed across candidates.
Define
\begin{equation}
\begin{aligned}
J(A)&=\sum_{i\in\mathcal P}\|\Phi_{\mathcal P,i}(A)-\bm z_{g,i}\|_2^2,\\
J_{\mathcal S}^{\mathrm{oracle}}(A)&=\sum_{i\in\mathcal S}\|\Phi_{\mathcal P,i}(A)-\bm z_{g,i}\|_2^2,\qquad
J_{\mathcal S}(A)=\sum_{i\in\mathcal S}\|\Phi_{\mathcal S,i}(A)-\bm z_{g,i}\|_2^2.
\end{aligned}
\label{eq:comparison_costs}
\end{equation}
The oracle cost retains the selected positions but uses full-token dynamics.
Set $D_{\mathcal S}=J-J_{\mathcal S}^{\mathrm{oracle}}$ (discarded cost) and $E_{\mathcal S}=J_{\mathcal S}-J_{\mathcal S}^{\mathrm{oracle}}$ (change from sparse dynamics).
Exactly,
\begin{equation}
J_{\mathcal S}(A)-J(A)=E_{\mathcal S}(A)-D_{\mathcal S}(A).
\label{eq:sparse_cost_decomposition}
\end{equation}
Sparse planning therefore changes both which positions enter the cost and the predictions at retained positions.
Only action-dependent variation in $E_{\mathcal S}-D_{\mathcal S}$ can change candidate rankings; an action-independent offset preserves them.
\end{revision}

\end{document}